\documentclass[nopreprintline,12pt]{elsarticle}

\usepackage{amssymb}
\usepackage{changepage}
\usepackage{longtable} 
\usepackage{booktabs}
\usepackage{placeins}
\usepackage{pifont} 
\usepackage{xcolor}
\usepackage[colorlinks]{hyperref}
\usepackage{comment}

\definecolor{darkgreen}{rgb}{0,0.5,0}
\usepackage[colorinlistoftodos]{todonotes}
\usepackage{caption}
\usepackage{subcaption}
\usepackage{pifont}

\usepackage{dirtytalk}

\usepackage{multirow}
\usepackage{amsmath}

\usepackage{soul}
\setulcolor{red} 
\setstcolor{red} 
\sethlcolor{yellow} 

\journal{Transportation Research Part C}

\begin{document}

\begin{frontmatter}

\title{Mesoscale Traffic Forecasting for Real-Time Bottleneck and Shockwave Prediction}

\affiliation[inst1]{organization={Department of Civil and Environmental Engineering, University of California, Berkeley},
            city={Berkeley},
            postcode={94720}, 
            state={CA},
            country={USA}}

\affiliation[inst3]{organization={Center for Robotics, Mines Paris, PSL University},
            addressline={60 boulevard Saint-Michel}, 
            city={Paris},
            postcode={75006}, 
            state={IDF},
            country={France}}

\affiliation[inst2]{organization={Valeo Driving Assistant Research},
            addressline={6 rue Daniel Costantini}, 
            city={Créteil},
            postcode={94000}, 
            state={IDF},
            country={France}}

\author[inst1,inst3,inst2]{Raphael Chekroun}
\author[inst1]{Han Wang}
\author[inst1]{Jonathan Lee}
\author[inst2]{Marin Toromanoff}
\author[inst3]{Sascha Hornauer}
\author[inst3]{Fabien Moutarde}
\author[inst1]{Maria Laura Delle Monache}

\begin{abstract}
Accurate real-time traffic state forecasting plays a pivotal role in traffic control research. In particular, the CIRCLES consortium\footnote{https://circles-consortium.github.io/} project necessitates predictive techniques to mitigate the impact of data source delays. After the success of the MegaVanderTest experiment \cite{Lee2024Megacontroller}, this paper aims at overcoming the current system limitations and develop a more suited approach to improve the real-time traffic state estimation \cite{wang2024speed} for the next iterations of the experiment. In this paper, we introduce the SA-LSTM, a deep forecasting method integrating Self-Attention (SA) on the spatial dimension with Long Short-Term Memory (LSTM) yielding state-of-the-art results in real-time mesoscale traffic forecasting. We extend this approach to multi-step forecasting with the $n$-step SA-LSTM, which outperforms traditional multiforms-step forecasting methods in the trade-off between short-term and long-term predictions, all while operating in real-time. 



\end{abstract}




\end{frontmatter}


\section{Introduction \& State of the Art}
\label{sec:Introduction}


Traffic forecasting stands as a pivotal research challenge in contemporary industrial academia. With the impending advent of autonomous vehicular systems, the imperative of accurate traffic prediction is accentuated, primarily due to its potential ramifications on urban design, public safety, and the overarching efficacy of transportation infrastructures. Anticipating forthcoming traffic conditions enables stakeholders—ranging from policymakers to urban strategists—to allocate resources judiciously, institute infrastructural enhancements in a timely manner, and conceptualize efficacious traffic governance methodologies. Such a proactive stance not only ameliorates congestion but also mitigates accident risks, attenuates environmental ramifications, and culminates in both temporal and financial savings for commuters and the broader populace. Traffic information is relevant on several levels of granularity, on a scale from micro to macro, each presenting its own interest. Micro-scale traffic information captures detailed, vehicle-level data, such as individual speeds, positions, and behaviors, providing a high-resolution view of traffic conditions at specific locations. It is often used for fine-grained analyses, like understanding the dynamics of a single intersection, or collaborative planning to enable an energy-efficient driving \cite{colab, delle2019feedback, stern2018dissipation}. On the other hand, macro-scale traffic information focuses on aggregated, high-level data that provides an overall picture of traffic flow across broader areas. This can include metrics like average speeds, traffic volumes, and congestion levels, and is generally employed for long-term planning and large-scale traffic management \cite{eta}. Mesoscale traffic information occupies the middle ground between micro-scale and macro-scale. Specifically in the studied use case, it focuses on how groups of vehicles interact with each others on segmented portions of a single highway and how it impact average speeds acrross these distincts sections. These three types of information offer valuable insights but differ in their level of detail and computational requirements.



Traffic forecasting has long been explored via rule-based methods. In particular, some research extended the Kalman Filter for traffic estimation via ensemble methods \cite{yuan2015efficient} or Kalman recursions in dynamic state-space \cite{portugais2014adaptive}. Alternative modelizations, such as particle filters \cite{ren2010freeway} or spatial copulas \cite{ma2019spatial}, have also been leveraged to this extent.
However, these methods suffer from performance decays when unexpected events provoke nonpredictable changes or if the allocation to a traffic pattern is inaccurate.

The advent of deep learning has addressed several shortcomings of rule-based methods. By learning from data, these models can account for unpredictable yet regular behaviors. Laña et al. \cite{lana2019adaptive} employed Spiking Neural Networks to achieve long-term pattern forecasting, adapting these predictions to real-time situations. For short-term forecasting, the Graph Convolution Network (GCN) has emerged as a potent tool. Guo et al. \cite{guo2020dynamic} utilized a GCN for traffic forecasting, integrating it with a latent network to glean spatial-temporal features. Mallick et al. \cite{mallick2022deep} enhanced the capabilities of GCN by incorporating ensembling methods, leveraging Bayesian hyperparameter optimization and generative modeling. However, despite their efficiency, these deep models consist of computationally demanding operations, making them unsuitable for real-time forecasting.





\begin{figure}[!ht]
    \centering
    \includegraphics[width=\linewidth]{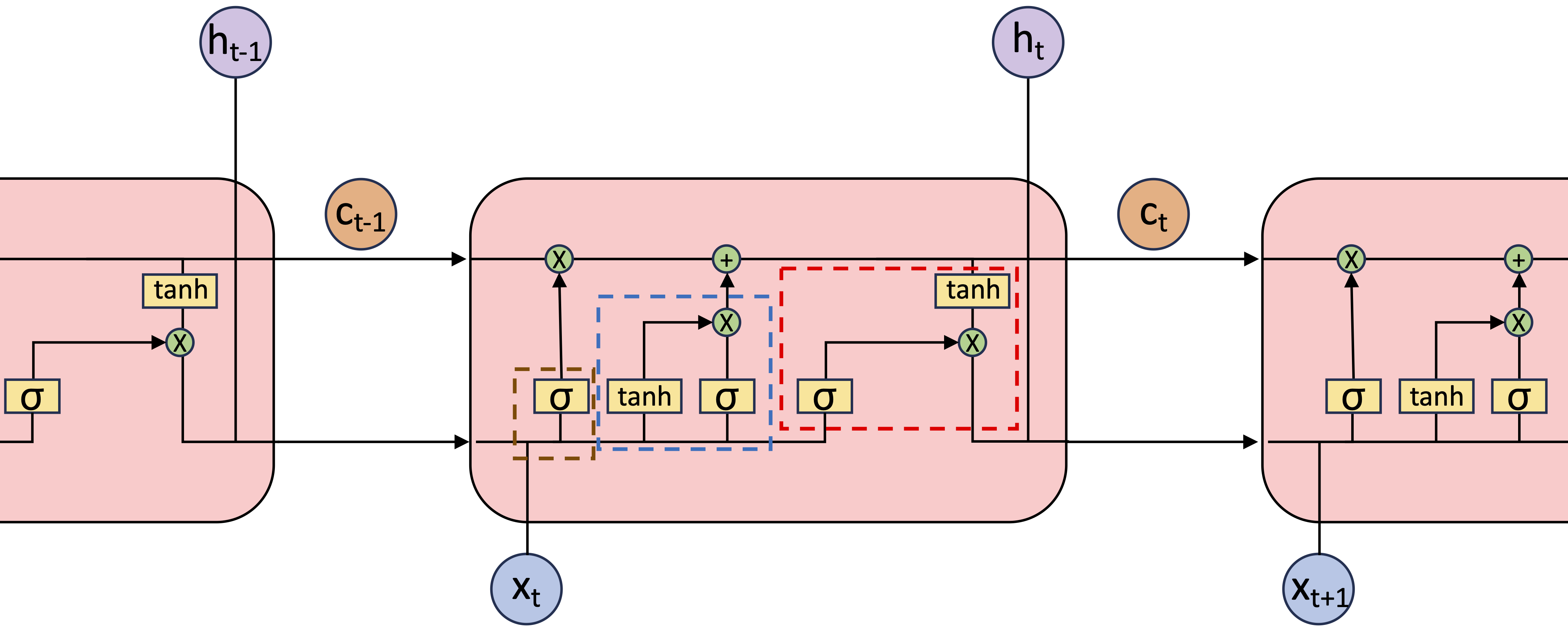}
    \caption{Representation of an LSTM cell. The Cell State ($C_t$, in orange) runs through the entire sequence. It stores and transmits information across time steps while selectively modifying or forgetting parts of it. The Hidden State ($h_t$, in purple) is the output of the LSTM cell at a specific time step. It carries information that is relevant to the current time step's prediction or output. It is also influenced by the cell state and the input at that time step. LSTMs employ three gate types (forget in brown, input in blue, and output in red) to regulate how information is managed within the cell state and the hidden state.}
    \label{fig:LSTM}
\end{figure}

Recurrent Neural Networks (RNN), and in particular Long Short Term Memory (LSTM) \cite{lstm}, are lighter deep-based methods for forecasting able to effectively to capture and model sequential data via a sophisticated memory mechanism. Key components of LSTM networks are represented in Figure \ref{fig:LSTM}. During training, the LSTM network learns to adjust the parameters of its gates and the cell state in a way that allows it to capture long-range dependencies and patterns in sequential data. This enables LSTMs to excel in time series prediction where understanding context and dependencies over time is crucial. However, LSTM remains limited to capturing both spatial and temporal patterns in series prediction. 

Incorporating an analysis of spatial dependencies has been explored through different methods. ConvLSTM replaces LSTM state-to-state and  input-to-state transition with convolutions\cite{convlstm, sa-convlstm} to bring LSTM the computational capability to analyze spatiotemporal series. Methods relying on attention are able to learn how each data point interacts with each other at each timestep. In particular self-attention has been successfully leveraged with LSTM for diverse forecasting tasks \cite{sa-1, sa-2, sa-3}, and Transformers \cite{transformers,transformers_ds_review} recently showed promising results in data series \cite{transformers-data-series}. Some methods leverages both convolution and self-attention to reach state-of-the-art results on some datasets \cite{sa-convlstm}. Other methods combine LSTM and graph neural network \cite{graph_survey,graph,graph_lstm} to forecast and quantify how roads and intersections impact one another through graph modelization. However, these methods are by design for macro-scale road systems and are unfit for meso modelizations. 








In this study, our attention is directed toward mesoscale traffic forecasting, which occupies the middle ground between micro-scale and macro-scale analysis. Specifically, we examine how groups of vehicles interact on segmented portions of a single highway. We aim to determine the average speed of vehicles across these distinct sections. The data is limited to highway conditions and does not incorporate information from entry or exit ramps. The ultimate goal is to forecast in real-time the development of traffic bottlenecks and shockwaves as part of the \textit{Congestion Impacts Reduction via Connected Autonomous Vehicles (CAV)-in-the-loop Lagrangian Energy Smoothing} (\href{https://circles-consortium.github.io/}{CIRCLES}) project, which seeks to mitigate traffic congestion and energy waste by utilizing Connected Autonomous Vehicles (CAV) on highways. 

This paper is organized as follows: In Section 2, we present the data source and how we modelize it as a data series problem. In Section 3 we present the methodology we developed for one-minute forecasting and multi-step forecasting. In Section 4, we present ablations studies and experimental results to justify our methods. Finally, Section 5 concludes this paper.

\section{Data Collection and Forecasting Methodology}
\label{sec:Data Collection and Forecasting Methodology}
\subsection{Data Acquisition}
\label{sec:data_inrix}
\begin{figure}
\centering
    \includegraphics[height=8cm]{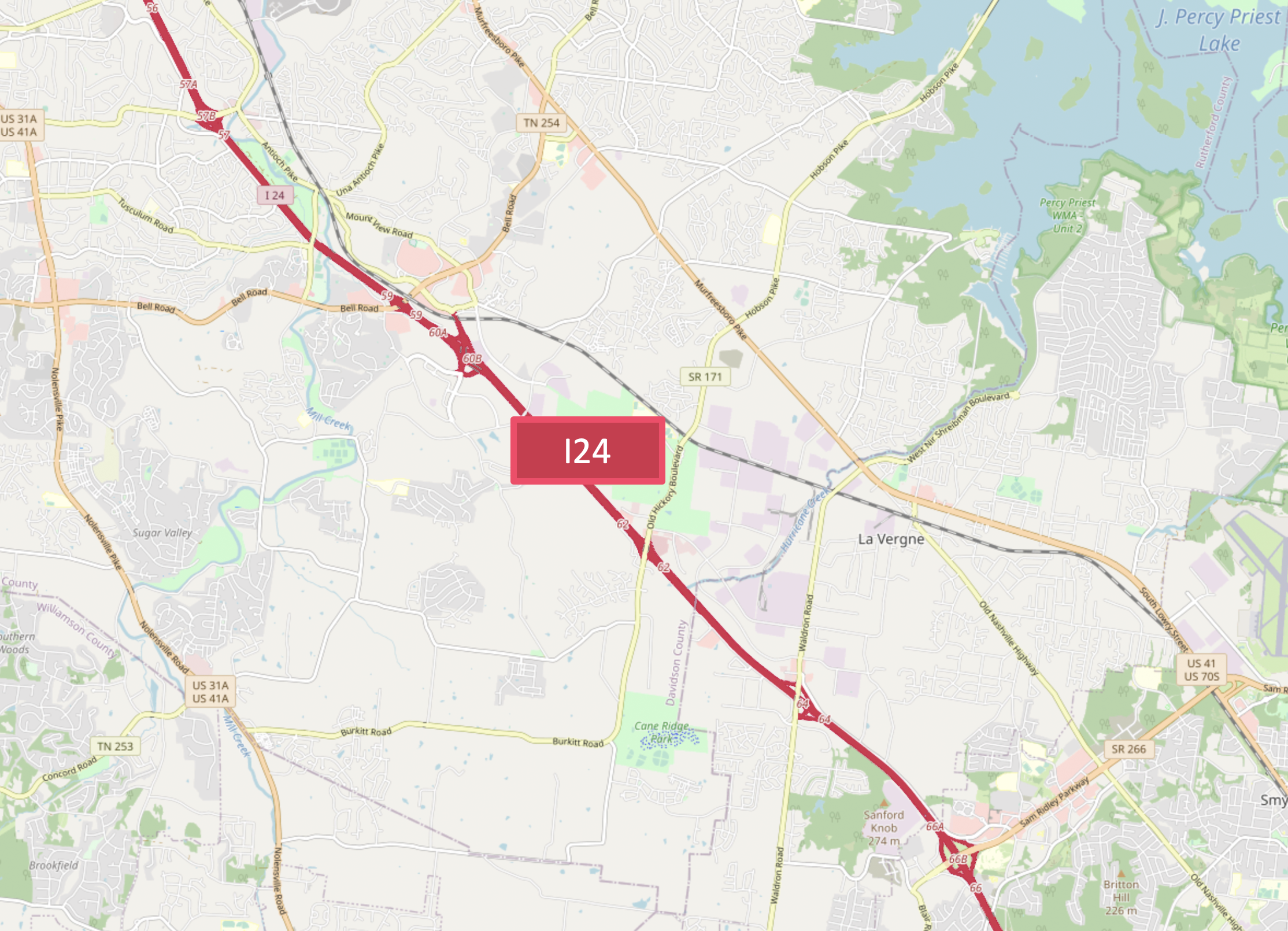}
    \caption{The Target Road Segment of CIRCLES: I-24 Westbound in Nashville, Tennessee, seen within the highlighted region.}
\label{fig:i24}
\end{figure}

This research utilizes mesoscale data obtained from INRIX traffic services \cite{inrix}, which includes average speeds across multiple lanes on 21 segments of the I-24 interstate highway in Nashville, TN, as depicted in Figure \ref{fig:i24}. 

\begin{figure}[h!]
    \centering
    \includegraphics[height=10cm]{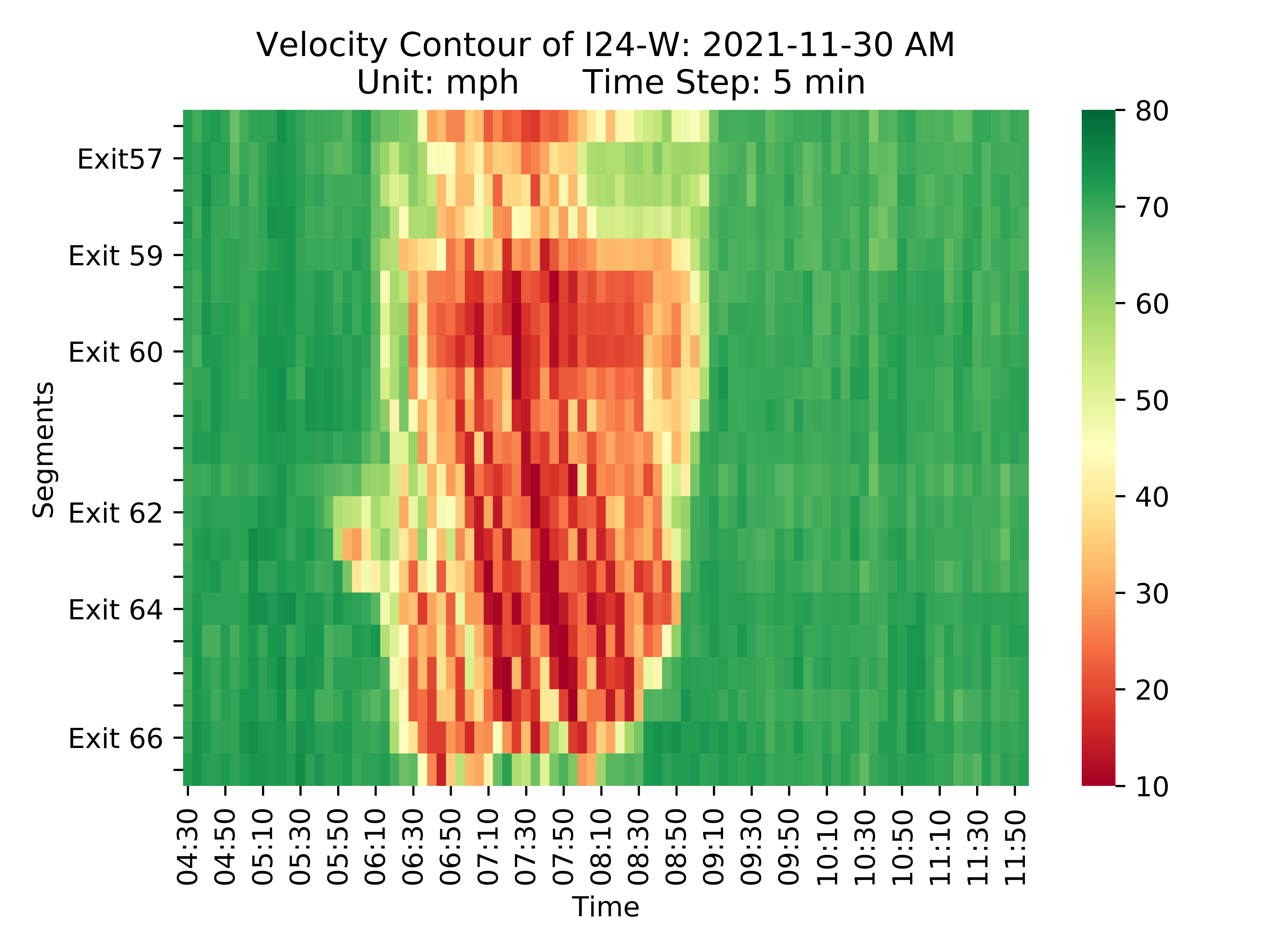}

    \caption{In the red contour of the figure, one observes the chronological progression of congestion on the specified segments. A notable persistent bottleneck is evident at Exit 59. This congestion initiates at approximately 6:00 a.m., likely attributable to the augmented commuting demand upstream, and it fully resolves by around 9:00 a.m.}
    
    \label{fig:inrix_demo}
\end{figure}

This data spans mileposts MM66 to MM59, covering an 11.4 km road fraction divided into 21 segments, with a sampling rate of 3,600 data points per day. An example of typical morning traffic is shown in Figure \ref{fig:inrix_demo}.

While INRIX traffic data updates every minute, there can be a slight lag of up to three minutes in data generation. The focus of this study is to enhance the accuracy and timeliness of traffic forecasting, especially considering the brief delays in data updates. The objective is to develop a predictive model that effectively forecasts traffic patterns in three-minute intervals, leveraging the minute-by-minute data refreshment to anticipate and manage traffic conditions more efficiently.

\subsection{Modelization as a data series problem}
We modeled this data series problem in the following way. At every time-step $t$, we note $v^i_t$ the average velocity over the lanes on the whole $i\in [0,20]$ segment. Hence, the studied data series can be seen as follows:
$$
\mathrm{V}_{t}=\left[\begin{array}{c}
\mathrm{v}_{t}^0 \\
\mathrm{v}_{t}^1 \\
\vdots \\
\mathrm{v}_{t}^{\mathrm{20}}
\end{array}\right],
$$

We also define
        $$ 
        \forall (t, k) \in \mathbb{N}^2,
        \mathrm{I}_{t}^{k}=V_t\oplus \cdots \oplus V_{t+k} = \left[\begin{array}{ccc}
            \mathrm{v}_{t}^0 & \dots & \mathrm{v}_{t+k}^0 \\
            \mathrm{v}_{t}^1 & \dots & \mathrm{v}_{t+k}^1 \\
            \vdots & &  \vdots \\
            \mathrm{v}_{t}^{\mathrm{20}} & \dots & \mathrm{v}_{t+k}^{\mathrm{20}}
        \end{array}\right]
        $$
The concatenation of $k$ consecutive velocity vector starting at time $t$.

Therefore, our final model should be fed with $I_{t-s}^s$ to output $I_{t}^3$, $s$ being the chosen sequence length used as input.

\subsection{Training and validation datasets}
\begin{figure}
\centering
    \includegraphics[width=\linewidth]{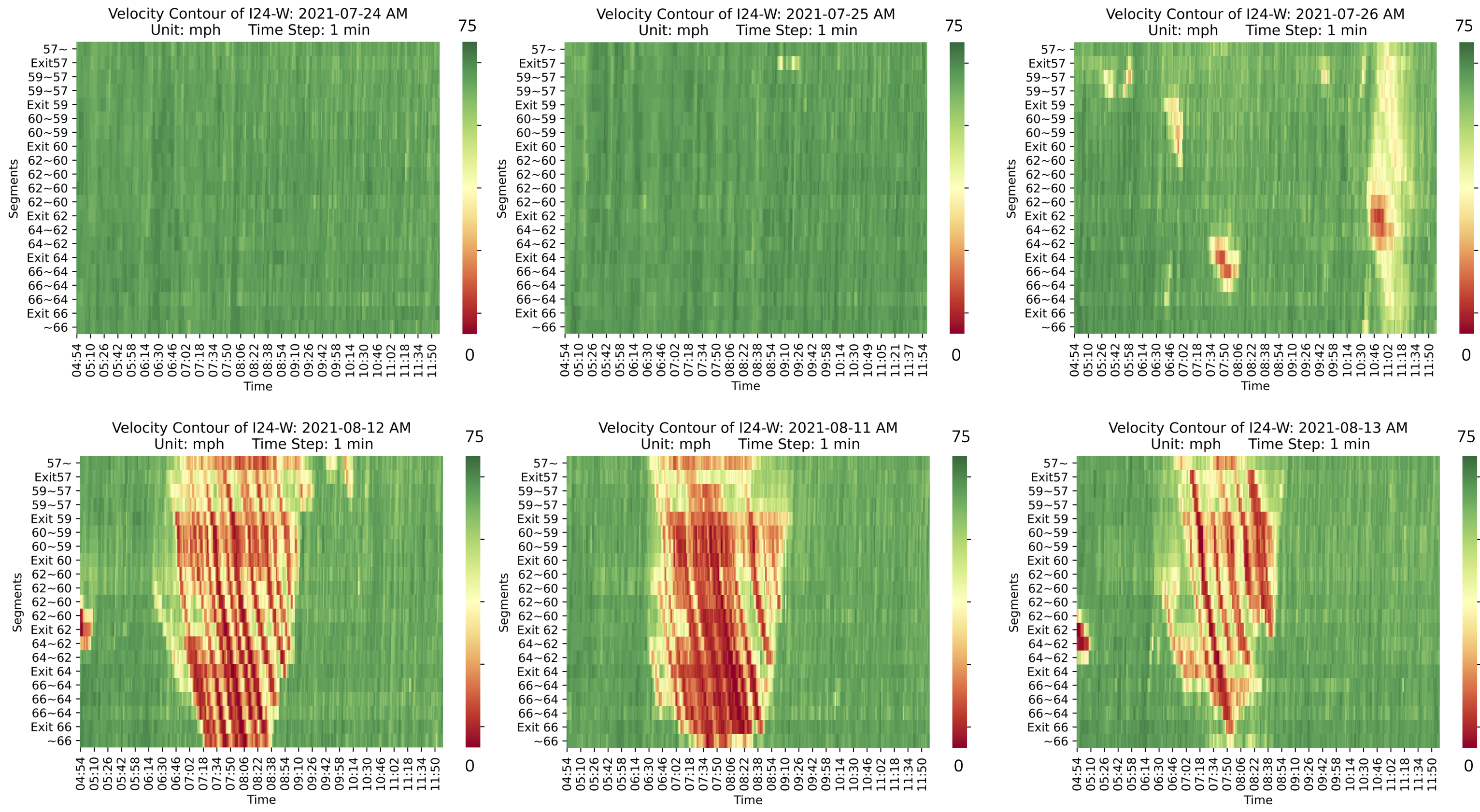}
\caption{Illustrative representation of the validation datasets. \textbf{Top Row:} Three representative snapshots from the Easy Validation set, showcasing common traffic patterns with periodic congestion and the prominence of temporal dependencies. \textbf{Bottom Row:} Three exemplar visuals from the Hard Validation set, highlighting moments of intense congestion, significant vehicle interactions, and the emphasis on spatial dependencies.}

\label{fig:validation_sets}
\end{figure}

The training set is composed of 504,000 data points (every minute for 350 days). We also built two validation sets, also represented in Figure \ref{fig:validation_sets}:

\begin{itemize}
    \item The Easy Validation set: made of 86,400 data points (every minute for 60 days), it mostly represents common traffic as most of it is smooth, with some discrete congestion and traffic shockwaves. There is usually at least one bigger traffic bottleneck every day between 6 am and 10 am. Traffic is mostly fluid in this dataset, interactions between vehicles are almost negligible, and metrics mostly represent a model capability to capture temporal dependencies. 

    \item The Hard Validation set: the composition of four 440 minutes of highly congested traffic bottlenecks (1,760 total data points). Metrics are evaluated independently and averaged on those three sets to obtain the Hard metric. Traffic being highly congested, interactions between vehicles are consequent, and validations metrics on this dataset represent a model capability to capture spatial dependencies. 
\end{itemize}

\section{Real-Time Mesoscale Traffic Forecasting}
\label{sec:traffic_forecasting}
Our primary emphasis is on single-step traffic prediction, which involves forecasting traffic conditions just one minute ahead. Subsequently, we expand our approach to solve the problem of multi-step traffic forecasting, which involves predicting traffic conditions several minutes into the future.

\subsection{One-minute INRIX Prediction}
\label{sec:one_minute}
While LSTM is a correct baseline for both accuracy and inference time, they do not qualify as an optimal solution for our data. Indeed, at a given time $t$, a traffic bottleneck at position $k$ will impact both the short and long-term $v_t^k$, but also the neighboring segments $v_t^{k-\epsilon}$ and $v_t^{k+\epsilon}$. Hence, studied data presents spatio-temporal relationships, while standard LSTM mostly focuses on temporal relationships. To overcome this limitations, self-attention can be a powerful tool. Indeed, self-attention can intuitively capture the dynamic dependencies between different segments of the road network, recognizing how traffic conditions on one segment affect others. By attending to relevant spatial and temporal patterns, self-attention enables traffic forecasting models to adapt and predict congestion, flow changes, and bottlenecks. This intuitive capacity to capture inter-dependencies makes self-attention a valuable asset in improving the accuracy and reliability of mesoscale traffic forecasting, ultimately contributing to more effective traffic management strategies and reduced congestion. Mathematically, self-attention update the tokens via a weighted by $X$ sum, with $X$ computed via Equation \ref{eq:attention}.

 \begin{equation}
    X = \text{softmax}(\frac{Q.K^T}{\sqrt{d}}) V
    \label{eq:attention}
\end{equation}

with the queries $Q$, keys $K$, and values $V$ being three tensors created through linear projection from the input tensor, and $d$ the feature size.

\begin{figure}[!h]
    \centering
    \includegraphics[width=\linewidth]{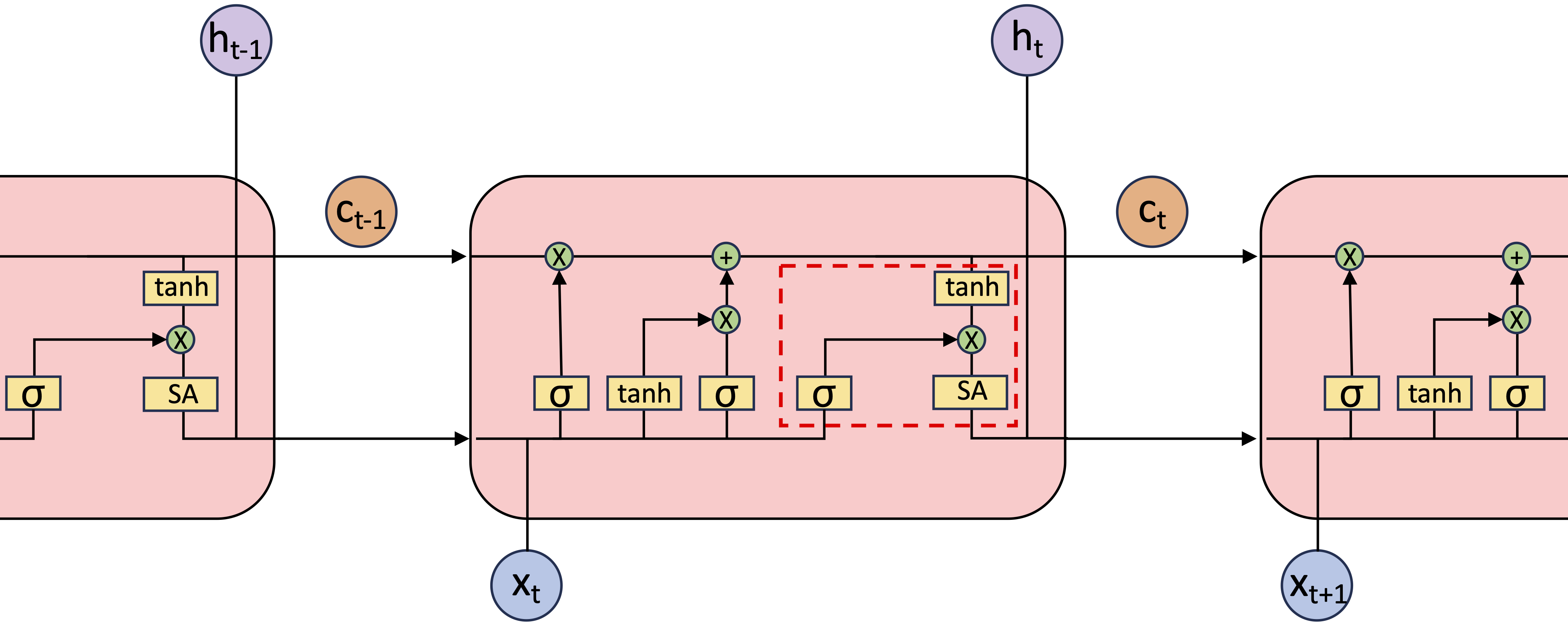}
    \caption{A single cell from an SA-LSTM network. The SA-LSTM is an LSTM in which the output gate, in red, is augmented with self-attention.}
    \label{fig:sa-lstm}
\end{figure}

Therefore, we designed a Self-Attention LSTM (SA-LSTM) whose output gate is augmented with a self-attention layer on the spatial dimension. Our SA-LSTM is represented in Figure \ref{fig:sa-lstm}.

To train the network to focus on more fine-grained spatial information without further increasing the computational time of operations at inference, we leveraged the Laplacian Pyramid loss \cite{laplacian_pyramid} mathematically defined in Equation \ref{eq:pyramid}.
\begin{equation}
    \mbox{Lap}_n\left(x, x^{\prime}\right)=\sum_{j=0}^n 2^{2 j}\left|L^j(x)-L^j\left(x^{\prime}\right)\right|_1
    \label{eq:pyramid}
\end{equation} 
where $L^j(x)$ is the j-th level of the Laplacian pyramid representation of $x$ \cite{laplacian}. It is a convolution-based loss able to weights the details at fine scales by capturing multi-scale information. It is used in addition to the MSE loss generally used for LSTMs.

\subsection{$n$-step SA-LSTM}
Section \ref{sec:one_minute} studied one-minute forecasting. In practice however, we aim to forecast up to three-minutes, \textit{i.e.} have access to  $\mathrm{V}_{t+1}, \mathrm{V}_{t+2},\mathrm{V}_{t+3}$. $n$-step forecasting is classically done via recursive inference over the data. Therefore, inferring $I_t^3$ is made in three successive inferences. First, the network is fed with $I_{t-s}^s$ and outputs $\tilde{v}^t$. Then, the network is fed with $I_{t-s+1}^s \oplus\tilde{v}^t$ to infer $\tilde{v}^{t+1}$, and so on. However, such methods suffer from accumulation error, as inaccuracies in each inference will weigh on the next ones. Also, total inference time is at least $n$ times the inference time of a single network inference. This method is therefore unfit for real-time inference. Another method is the all-at-once technique, in which a single LSTM is fed $I_{t-s}^s$ and trained to output $I_{t}^{t+n}$. While significantly faster and offering better long-term forecasting, this method can lead to underperformance on short term forecasting compared to 1-step LSTM which is not desirable in our case of study. 

\begin{figure}[!h]
    \centering
    \includegraphics[width=1\linewidth]{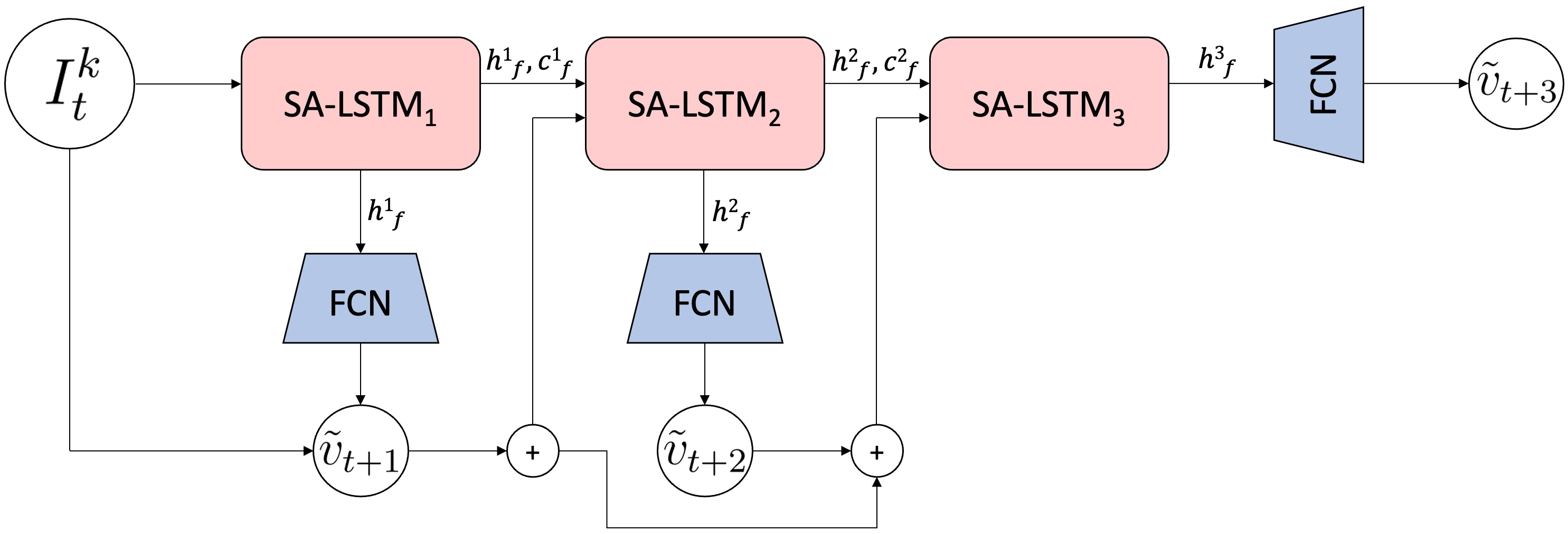}
    \caption{Each $i$-th layer of the $3$-step SA-LSTM is trained to infer the forecasting at time $t+i$  through a shared weight fully connected network (FCN). We note $h_f^i$ and $c_f^i$ the outputs of the last cell of the $i$-th SA-LSTM.}
    \label{fig:nstep_lstm}
\end{figure}

We designed the $n$-step SA-LSTM, a highly supervised multi-layer SA-LSTM represented in Figure \ref{fig:nstep_lstm}, to take the best of both world: a fast method resilient to accumulation error offering good short term and long term forecasting.

An $n$-step SA-LSTM is a $n$ layers SA-LSTM where:
\begin{itemize}
    \item Each layer output is constrained via a loss to converge toward $V_{t+i}$;
    \item Each $i$-th layer input is the concatenation of the network input $v_{t-k-1}...v_{t}$ concatenated with previous layer output $(\tilde{v}_{t+j})_{j\leq i}$. Layer $i$ also takes $h_{i-1}, c_{i-1}$ as input.
\end{itemize}  

Therefore, each layer have the same input and output dimensions but contains a different number of cells - which is equal to the input sequence length. Indeed, if input sequence length is 8, first layer will have 8 cells, second layer 8 + 1 cells as we add the previous layer output, and so on. The training of a $n$-step LSTM is sequential layer-wise as each layer is trained independently until convergence. 
\begin{itemize}
    \item Epochs $0 \rightarrow N$: $\mbox{layer}_1$ is trained, other layers are frozen and loss is only on $\tilde{v}_{t+1}$
    \item  Epochs $N \rightarrow 2\times N$: $\mbox{layer}_2$ is trained, other layers are frozen and loss is only on $\tilde{v}_{t+1}$
    
    \begin{center}
        $\vdots$
    \end{center}

    \item Epochs $(n-1)\times N \rightarrow n\times N$: all layers are unfrozen and the network is fine-tuned.
\end{itemize}

\section{Experimental results}
Unless specified otherwise, all presented models have been trained with an AdamW \cite{adamw} optimizer set with a learning rate of 0.01 and a scheduler to make the learning rate decrease by a factor of 10 when validation metrics stagnate or increase over 3 consecutive validations. Training aims to minimize the Mean Square Error (MSE) between the prediction $\tilde{y}$ and the ground truth $y$ \textit{i.e.}, the value $\frac{1}{n} \sum_{i=1}^n (y_i - \tilde{y_i})^2$. For ablation studies, training seeds are fixed, and gradient descent is not stochastic: every batch contains the whole dataset and hence is an epoch. 
\subsection{One-minute forecasting}

\textbf{Ablation study over self-attention.}
Comparison between LSTM and SA-LSTM is presented in Table \ref{Tab:vanille_ablation}. We observe that LSTM and SA-LSTM are on-par on the Easy validation set, and SA-LSTM is significantly better on the Hard dataset. Hence, adding a self-attention layer to an LSTM allows for enhancing the quality of spatial dependencies predictions with no degradation of temporal dependencies.

\begin{table}[!htb]
\centering
\begin{tabular}{cccc|c} 
\toprule
& & \multicolumn{2}{c}{Validation set} &  \\ \cmidrule{3-4}  
Method & Self-Attention & Easy & Hard & Time (ms) \\
\midrule 
\multirow{2}{*}{LSTM} & - & 0.66 & 5.71 & 0.2 \\
 & \ding{51} & 0.64 & \textbf{4.53} & 0.5\\
\bottomrule  
\end{tabular}
\caption{Ablation study of LSTM and SA-LSTM on INRIX data for traffic forecasting. Metric is MSE scaled by $\times 10^{3}$. Time is inference time measured as the mean over 50,000 inferences after a warmup of 1,000 inferences.}
\label{Tab:vanille_ablation}
\end{table}

\textbf{Ablation study over Laplacian Pyramid loss.}
Experimental results are presented in Table \ref{Tab:lap_ablation}. Training with this loss gave better results on one-minute predictions, particularly on the Hard dataset with a significant observed improvement. Indeed, this loss allows the model to focus the training on high-frequency details, which are more important in the Hard set. We observed the optimal depth to be 3 for our chosen hyper-parameters setting. Accuracy degrades for deeper depth than 3 because of the required pre-processing on tensors used for the Laplacian Pyramid loss during training only: we need the dimension of the inputs of this loss to be a multiple of $2^{\mbox{\small{depth}}}$, and going to deep adds substantial empty padding. Therefore most of the input becomes 0, which improves training metrics but significantly decreases validation metrics.

\begin{table}[!h]
\centering
\begin{tabular}{cccccccc} 
\toprule
& & \multicolumn{4}{c}{Pyramid depth} \\ \cmidrule{3-7} 
Method & Validation & 0 & 1  & 2  & 3 & 4 \\
\midrule 
\multirow{2}{*}{SA-LSTM} & Easy & 0.64 & 0.64 & \textbf{0.63} & \textbf{0.63} & 0.65 \\
& Hard & 4.48 & 4.31 & 3.94 & \textbf{3.59} & 4.12 \\
\bottomrule  
\end{tabular}
\caption{Ablation study of SA-LSTM trained with Laplacian Pyramid Loss using several depths on INRIX data for traffic forecasting. Metric is MSE and scaled by $\times 10^{-3}$. Inference time is unchanged compared to SA-LSTM, as the core network is the same. Next experiments will fix the Laplacian Pyramid loss depth at 3.}
\label{Tab:lap_ablation}
\end{table}

\textbf{Comparison with state-of-the-art methods.}
To validate our method, we compare inference time and validations accuracy with state-of-the-art spatio-temporal forecasting methods in Table \ref{Tab:final_ablation}. ConvLSTM \cite{convlstm} is a type of LSTM in which state-to-state and input-to-state transitions are replaced with convolutions. Self-Attention ConvLSTM \cite{sa-convlstm} is a ConvLSTM whose transitions have been augmented with self-attention layers. Transformers \cite{transformers} leverage attention to transform an input sequence into an output one by weighting how each elements of the input sequence interact one with each other. Interestingly, convolution-based methods trained with the $\mbox{Lap}_3$ loss led to a drop in accuracy on the easy validation set, while self-attention only methods experimentally benefit from it. SA-LSTM yields the best metrics on the Hard validation set and is comparable with the best method on the Easy one. Moreover, inference time is significantly lower than other methods designed for spatio-temporal forecasting and stays well under the intended millisecond. 

\begin{table}[h!]
\centering
\begin{tabular}{cccc|c} 
\toprule
& & \multicolumn{2}{c}{Validation Set} \\ \cmidrule{3-4} 
Method & $\mbox{Lap}_3$ Loss & Easy & Hard & Time (ms) \\
\midrule 
\multirow{2}{*}{LSTM} & \ding{55} & 0.66 & 5.71 & \multirow{2}{*}{\underline{\textbf{0.2}}} \\
 & \ding{51} & \textbf{0.61} & 4.09 &  \\
\midrule  
\multirow{2}{*}{ConvLSTM} & \ding{55} & 0.63 & 5.15 & \multirow{2}{*}{3.7} \\
 & \ding{51} & 0.68 & 5.13 &  \\
\midrule  
\multirow{2}{*}{SA-ConvLSTM} &  \ding{55} & 0.72 & 4.19 & \multirow{2}{*}{4.1}  \\
 &  \ding{51} & 0.76 & 3.94 &  \\
\midrule  
\multirow{2}{*}{Transformers} &  \ding{55} & 0.65 & 5.03 & \multirow{2}{*}{1.8} \\
&  \ding{51} & 0.64 & 4.71 &  \\
\midrule  
\multirow{2}{*}{\textbf{SA-LSTM}} & \ding{55} & 0.64 & 4.52 & \multirow{2}{*}{\underline{0.5}} \\
 & \ding{51} & 0.63 & \textbf{3.58} &  \\
\bottomrule  
\end{tabular}
\caption{Comparison of different forecasting methods and ablation study over the $\mbox{Lap}_3$ loss on INRIX data for traffic forecasting. Metric is MSE scaled by $\times 10^{3}$.}
\label{Tab:final_ablation}
\end{table}

Notably, both ConvLSTM and SA-ConvLSTM results on the Easy dataset degrade when training with a Laplacian Pyramid Loss but improve on the Hard dataset, as seen in the ablation in Table \ref{Tab:final_ablation}.
More generally, we observed the Laplacian Pyramid Loss to improve all methods on the Hard validation dataset, however, SA-LSTM trained with Laplacian Pyramid Loss still outperforms other variations.
Our intuition is that ConvLSTM-based models are by design highly focused on spatial dependencies and less on temporal ones than regular LSTM-based methods. Training with this loss worsens the spatial-dependency/ temporal-dependency analysis trade-off and over-advantages the analysis of spatial predictions over temporal ones.  

\begin{figure}[!ht]
    \centering
    \includegraphics[width=0.75\linewidth]{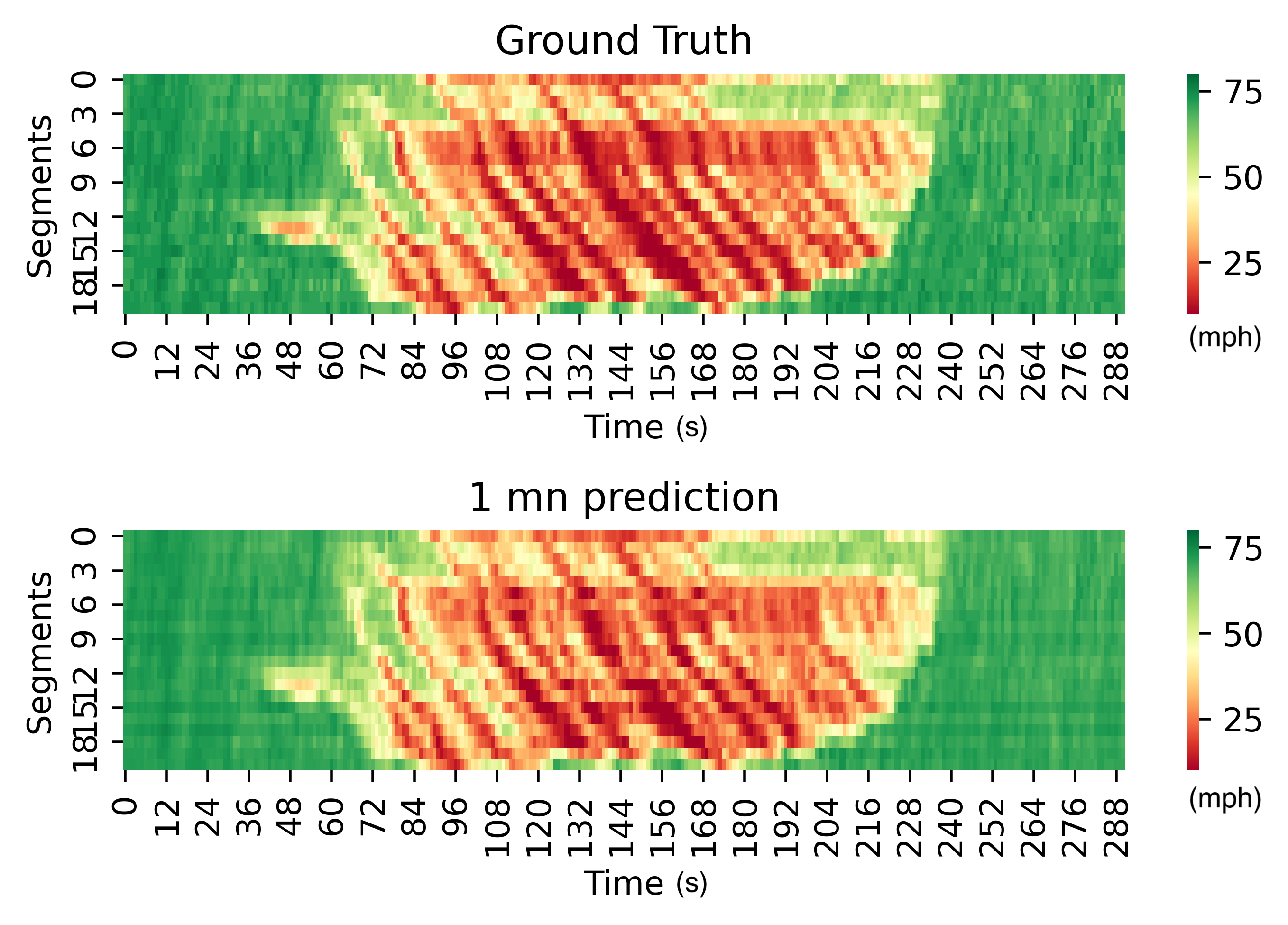}
    \caption{Comparison of heatmap generated from $\mbox{Lap}_3$ SA-LSTM one-minute traffic prediction with an heatmap generated from ground truth data. We observe inference to be as expected in both fluid and congested setup. Speeds are in miles/hour, time is in minute.}
    \label{fig:heatmap_sa_lstm}
\end{figure}

\begin{figure*}
    \begin{subfigure}[t]{.5\textwidth}
      \centering
        \includegraphics[width=1.0\linewidth]{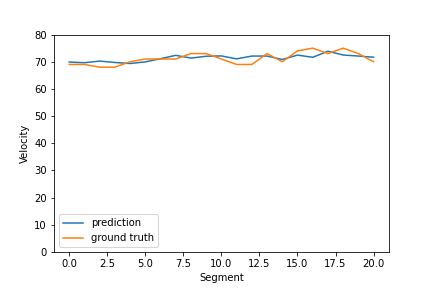}  \caption{Timestep 24: During the free flow stage, the prediction is able to generate the results around the free flow speed, 70 mile/hr.}
    \end{subfigure}%
    \begin{subfigure}[t]{.5\textwidth}
      \centering
    \includegraphics[width=1.0\linewidth]{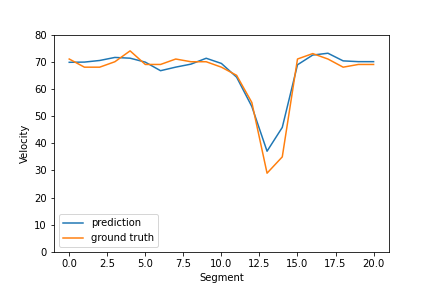}
      \caption{Timestep 48: A bottleneck start to form between segments 11 - 15. The prediction presents the same velocity change pattern with an accurate spatial location of the bottleneck as ground truth.}
    \end{subfigure}
    \begin{subfigure}[t]{.5\textwidth}
      \centering
        \includegraphics[width=1.0\linewidth]{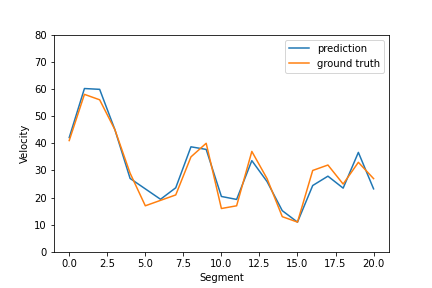}  \caption{Timestep 180: During the fully congested stage, the model is able to predict the propagation of the upstream shockwaves.}
    \end{subfigure}%
    \begin{subfigure}[t]{.5\textwidth}
      \centering
        \includegraphics[width=1.0\linewidth]{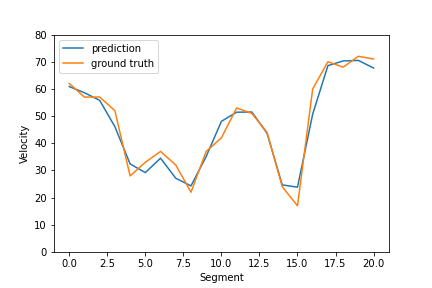}  \caption{Timestep 216: During the dissipation stage of the congestion, the prediction is able to capture the speed recovery at the bottleneck and upstream.}
    \end{subfigure}%
    \caption{Comparison between the ground truth and the one-minute predictions from the $\mbox{Lap}_3$ SA-LSTM during different stages of the congestion lifecycle. Velocities are in mph.}
    \label{fig:traffic_curve}
\end{figure*}

Overall, the model offering the best results on both the Easy and Hard datasets, so on temporal-focused and spatial-focused prediction, is the SA-LSTM. An example heatmap prediction from this network and corresponding traffic curve in different scenarios are represented in Figure \ref{fig:heatmap_sa_lstm} and Figure \ref{fig:traffic_curve}, in Section \ref{sec:qualitative_observations}.

\subsection{n-step forecasting}
We compared different multi step forecasting methods and compared metrics on $t+1$, $t+2$ and $t+3$. We also compare running time, as we want our solution to run under the millisecond.

\begin{table}[h!]
\centering
    \begin{tabular}{ccccc|c} 
    \toprule
     Method & Validation set & $t+1$ & $t+2$ & $t+3$ & Total time (ms)\\
    
    \midrule 
    
    \multirow{2}{*}{Recursive} & Easy & \textbf{0.63} & 0.83 & 1.24 & \multirow{2}{*}{1.8}\\
     & Hard & \textbf{3.58} & 6.51 & 10.67 & \\
    \midrule 
    \multirow{2}{*}{All-at-once} & Easy & 0.70 & \textbf{0.82} & \textbf{0.96} & \multirow{2}{*}{\underline{0.5}}\\
     & Hard & 4.31 & 5.58 & \textbf{7.43} & \\
     \midrule 
         \multirow{2}{*}{$n$-step} & Easy & \textbf{0.63} & 0.83 & 1.03 & \multirow{2}{*}{\underline{0.9}}\\
     & Hard & \textbf{3.58} & \textbf{5.41} & 7.56 & \\
     
     \bottomrule  

    \end{tabular}
         
    \caption{Comparison of different multi step forecasting methods. Metrics are MSE scaled by $10^{3}$. Underlined running times are the one acceptable for our application case.}
    \label{Tab:multi-step-forecasting}
\end{table}

The most optimal results for $t+1$ are achieved using the recursive and 3-step methods. This is expected since the LSTM weights dedicated to this inference were trained specifically for 1-step forecasting using real INRIX data. In contrast, the underperforming all-at-once method is not as finely tuned for 1-step predictions. However, from $t+2$ onwards, the accumulation errors begin to impact the recursive method, which then gets surpassed by both the all-at-once and 3-step approaches, leading to similar performance metrics. This gap becomes even more pronounced at $t+3$, where both the all-at-once and 3-step methods significantly outpace the recursive method. It's worth noting that the 3-step LSTM offers the best overall results for both single-step and multi-step predictions while maintaining sub-millisecond inference times. This highlights the method's proficiency in 1-step forecasting and its resistance to cumulative errors.

\begin{figure}[!ht]
    \centering
    \includegraphics[width=0.75\linewidth]{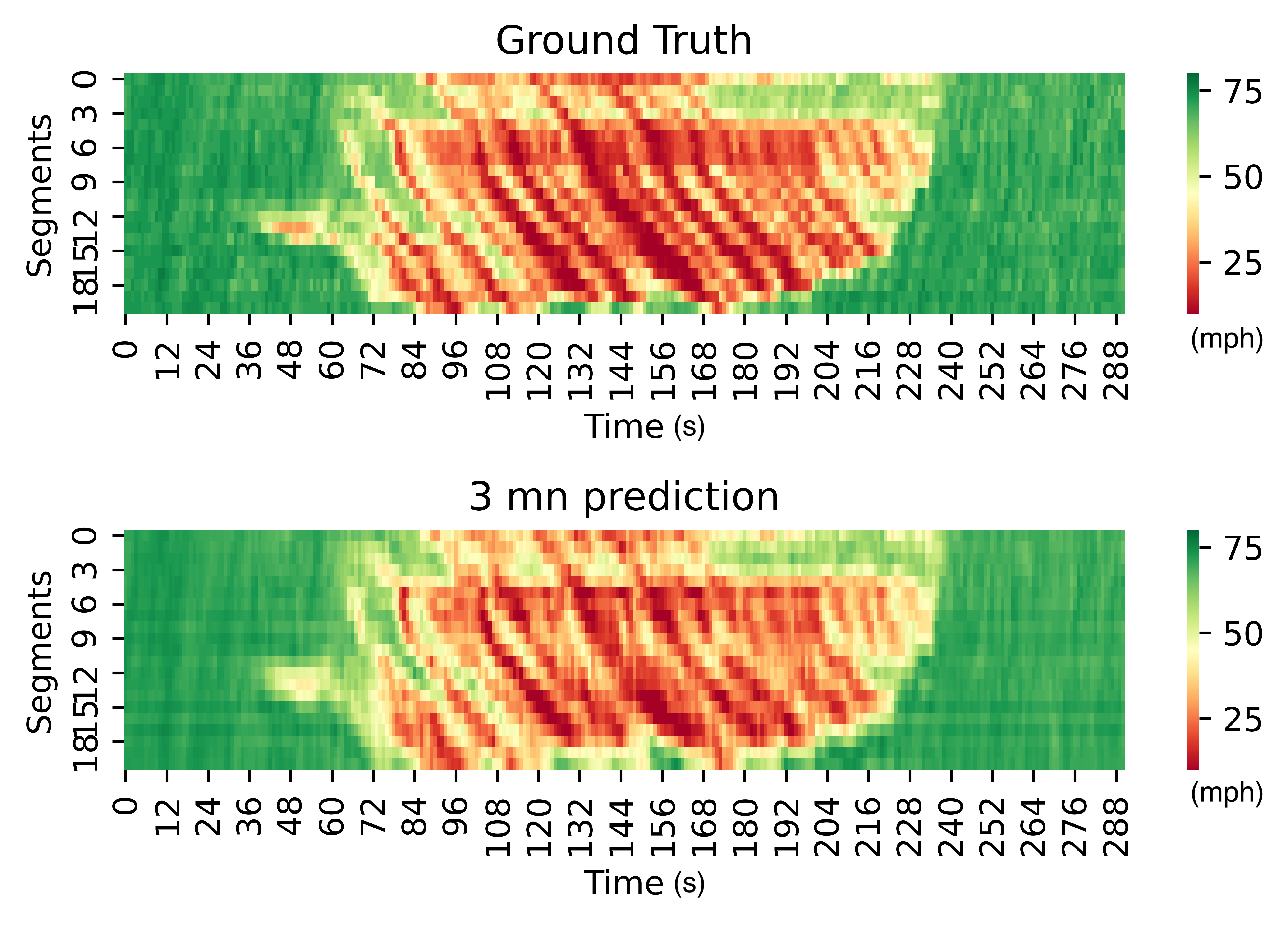}
    \caption{Comparison of heatmap generated from 3-step $\mbox{Lap}_3$ SA-LSTM three-minute traffic prediction with an heatmap generated from ground truth data. We observe inference to be as expected in both fluid and congested setup. Speeds are in miles/hour, time is in minute.}
    \label{fig:heatmap_multi_sa_lstm}
\end{figure}

\begin{figure*}
\begin{subfigure}[t]{.5\textwidth}
  \centering
    \includegraphics[width=1.0\linewidth]{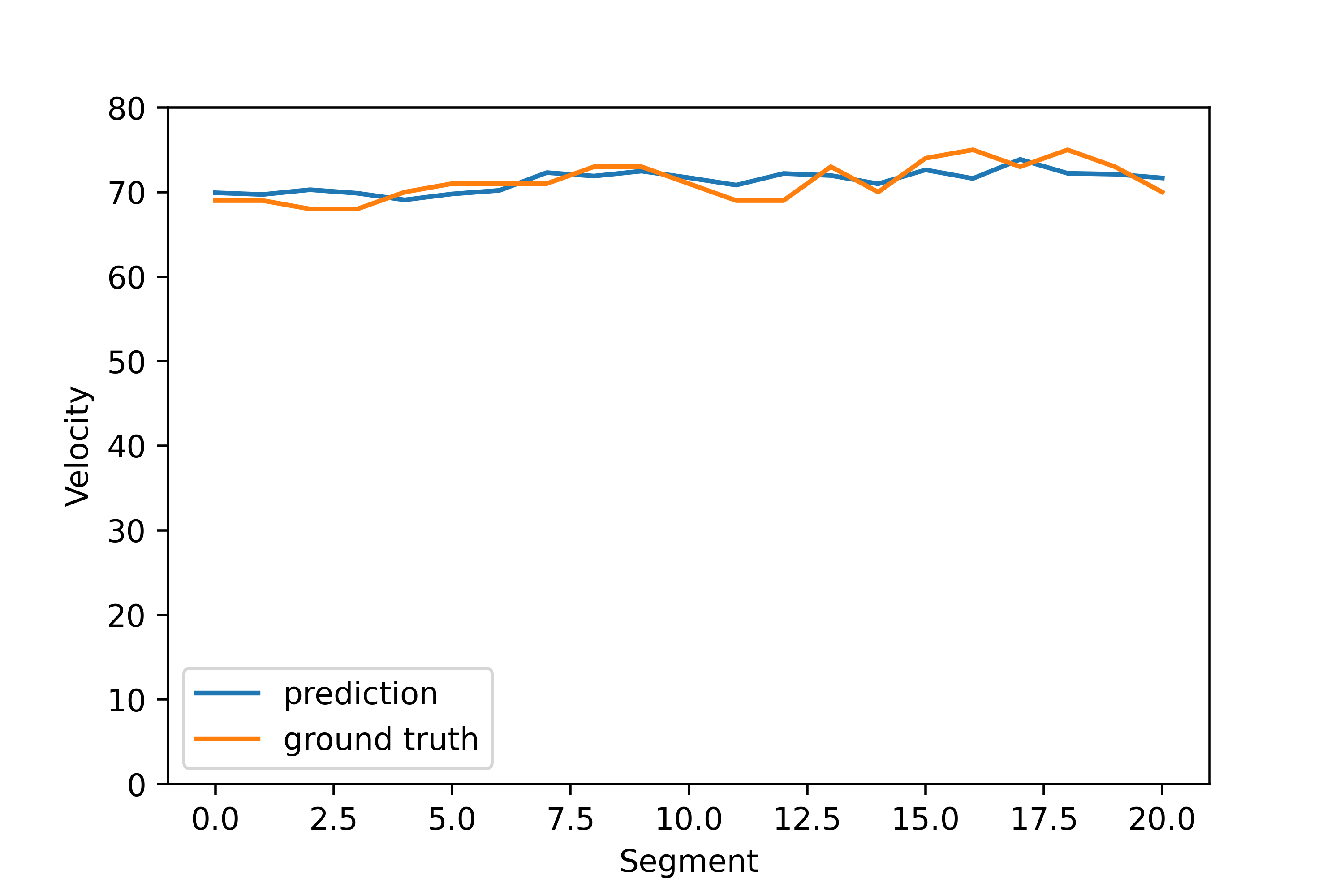}  
    \caption{Timestep 24: Prediction of free flow stage align with the ground truth around free flow speed.}
\end{subfigure}%
\begin{subfigure}[t]{.5\textwidth}
  \centering
\includegraphics[width=1.0\linewidth]{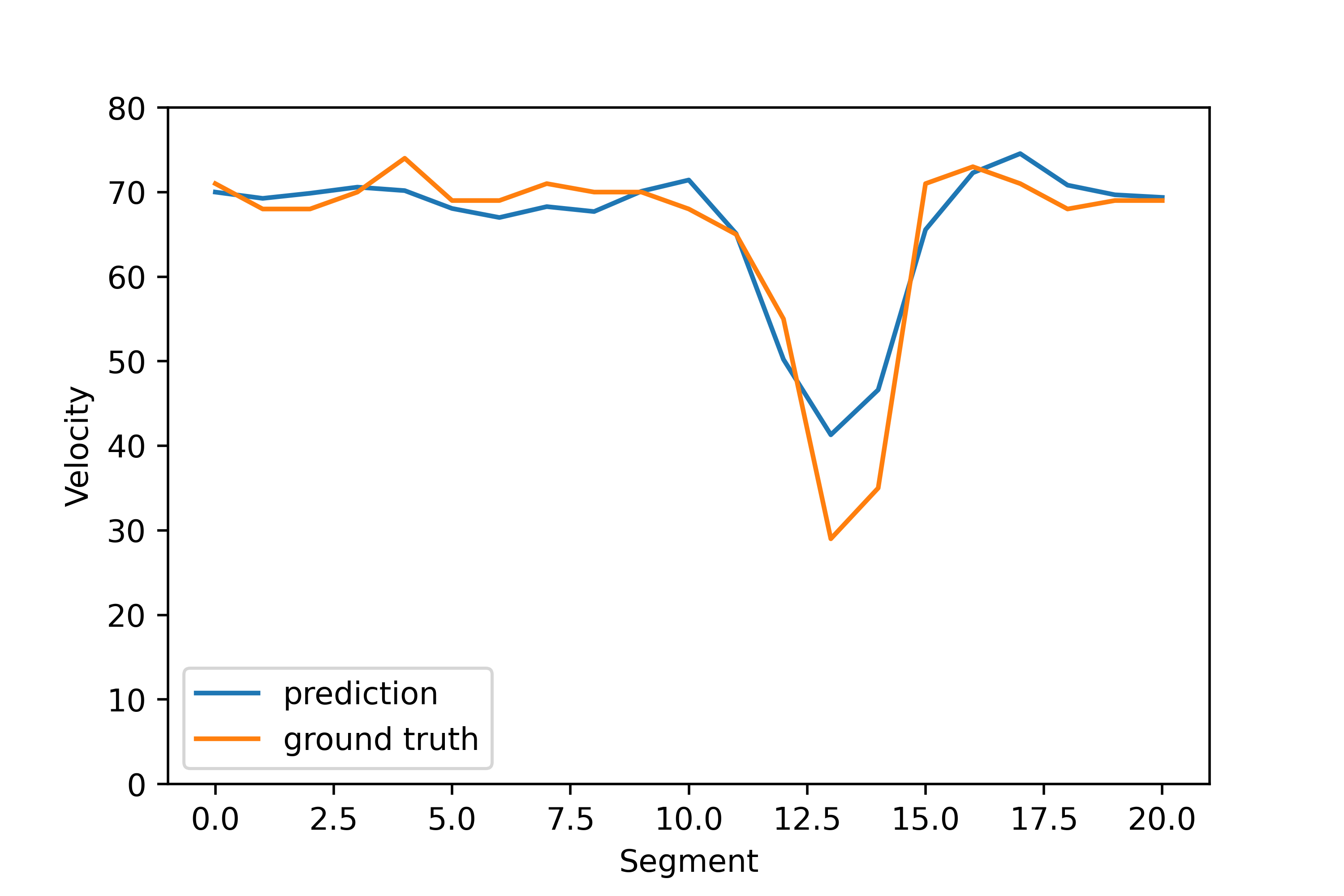}
  \caption{Timestep 48: A bottleneck start to form between segments 11 - 15. The prediction presents the same velocity change pattern with an accurate spatial location of the bottleneck as ground truth.}
\end{subfigure}
\begin{subfigure}[t]{.5\textwidth}
  \centering
    \includegraphics[width=1.0\linewidth]{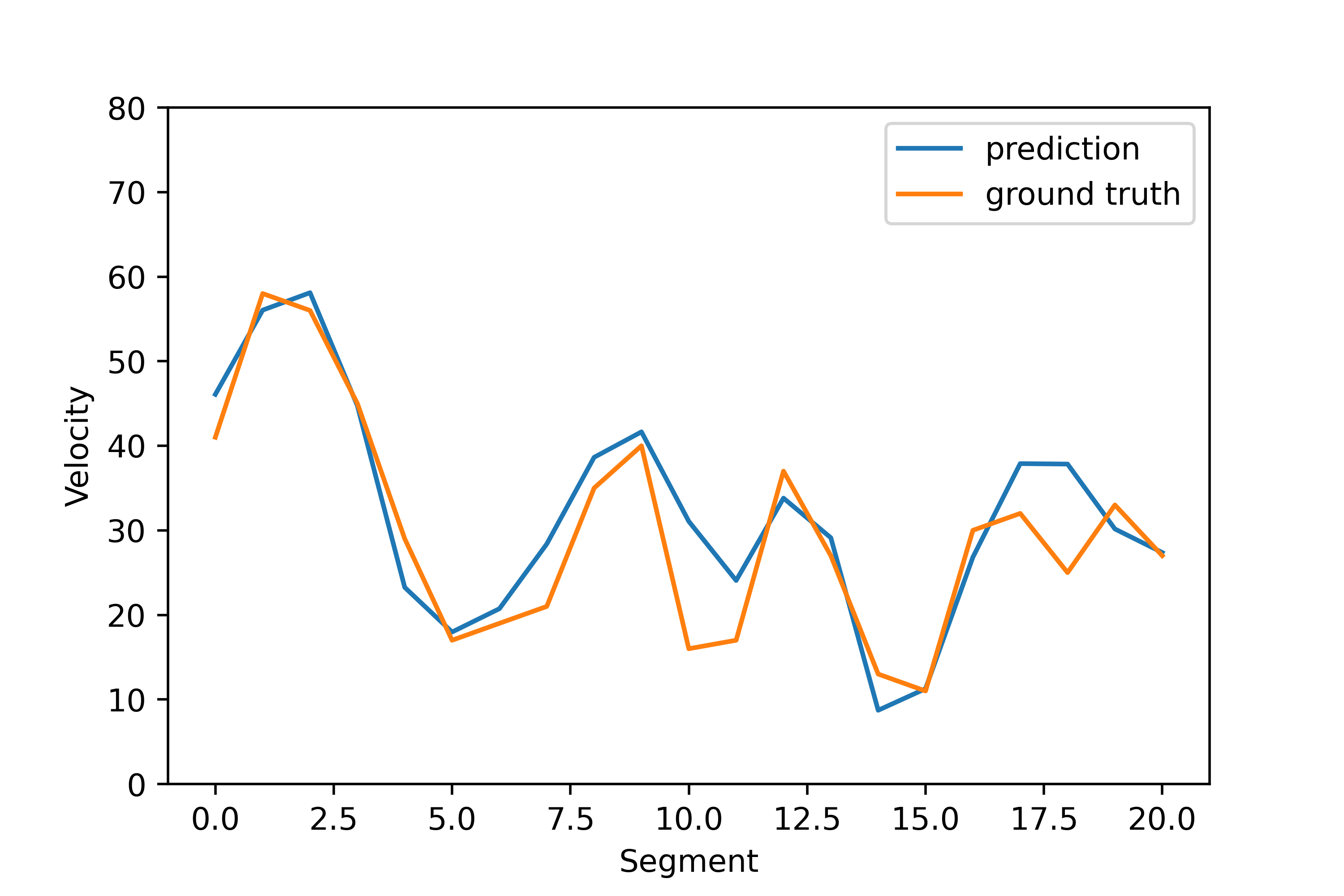}  
    \caption{Timestep 180: During the fully congested stage, the prediction captured the pattern of shockwave, while the prediction of absolute speed value has diversion from the ground truth. The predicted locations of the bottleneck and shockwaves are reliable.}
\end{subfigure}%
\begin{subfigure}[t]{.5\textwidth}
  \centering
    \includegraphics[width=1.0\linewidth]{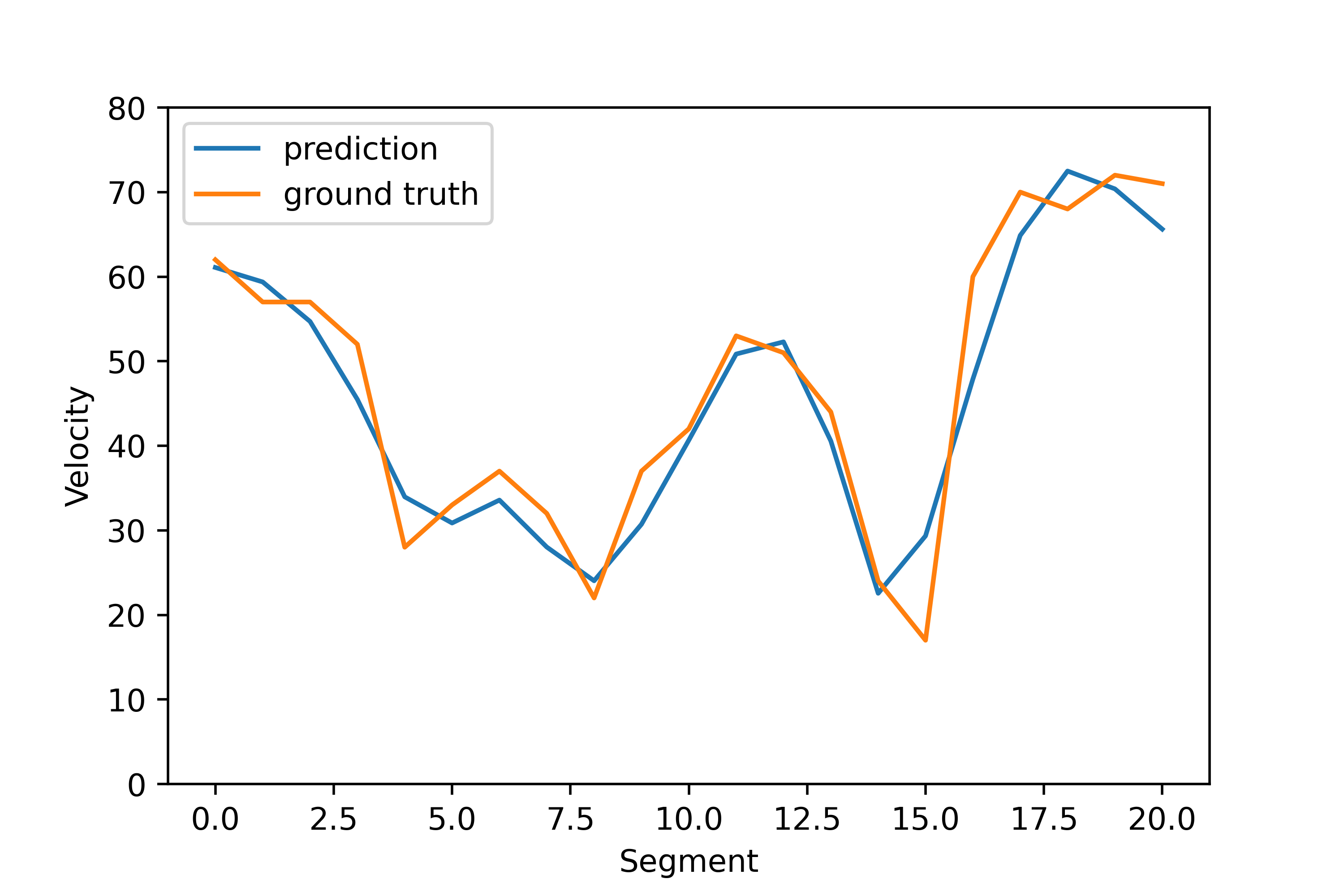}  
    \caption{Timestep 216: During the dissipation stage of the congestion, the prediction is able to capture the speed recovery at the bottleneck and upstream.}
\end{subfigure}%
\caption{Comparison between the ground truth and the three-minute predictions from the 3-step $\mbox{Lap}_3$ SA-LSTM during different stages of the congestion lifecycle. Velocities are in mph.}

\label{fig:multi_traffic_curve}
\end{figure*}

For our use case, $n$-step SA-LSTM appears as the best trade-off between inference time and both single and multi-step inference: $t+1$, $t+2$ and $t+3$ predictions are on-par with our best results overall while being faster than any other forecasting method except Vanilla LSTM.
An example heatmap prediction from this method and corresponding traffic curve in different scenarios are represented in Figure \ref{fig:heatmap_multi_sa_lstm} and Figure \ref{fig:multi_traffic_curve}.

\section{Qualitative observations}
\label{sec:qualitative_observations}
This section presents qualitative observations and analysis of some of our results. Presented qualitative representation comes in two forms of different granularity. 

\subsection{Heatmaps}
We first present a comparison of ground truth and inferred speed profile plotted has heatmaps for different forecasting methods and for both single step forecasting and multi-step forecasting. These heatmaps bear 3D information and represent mean velocity of all the vehicles on each segment of the studied part of the I-24 highway at each time step. While this kind of representation can give an overall insight on the quality of the inference, it is in practice hardly analyzable with the naked eyes. This subsection present heatmaps for single step $\mbox{Lap}_3$ SA-LSTM, 3-step $\mbox{Lap}_3$ SA-LSTM, and for the example of a failing case a 3 minute inference via the recursive method with the $\mbox{Lap}_3$ SA-LSTM.

\begin{figure}[!ht]
    \centering
    \includegraphics[width=0.75\linewidth]{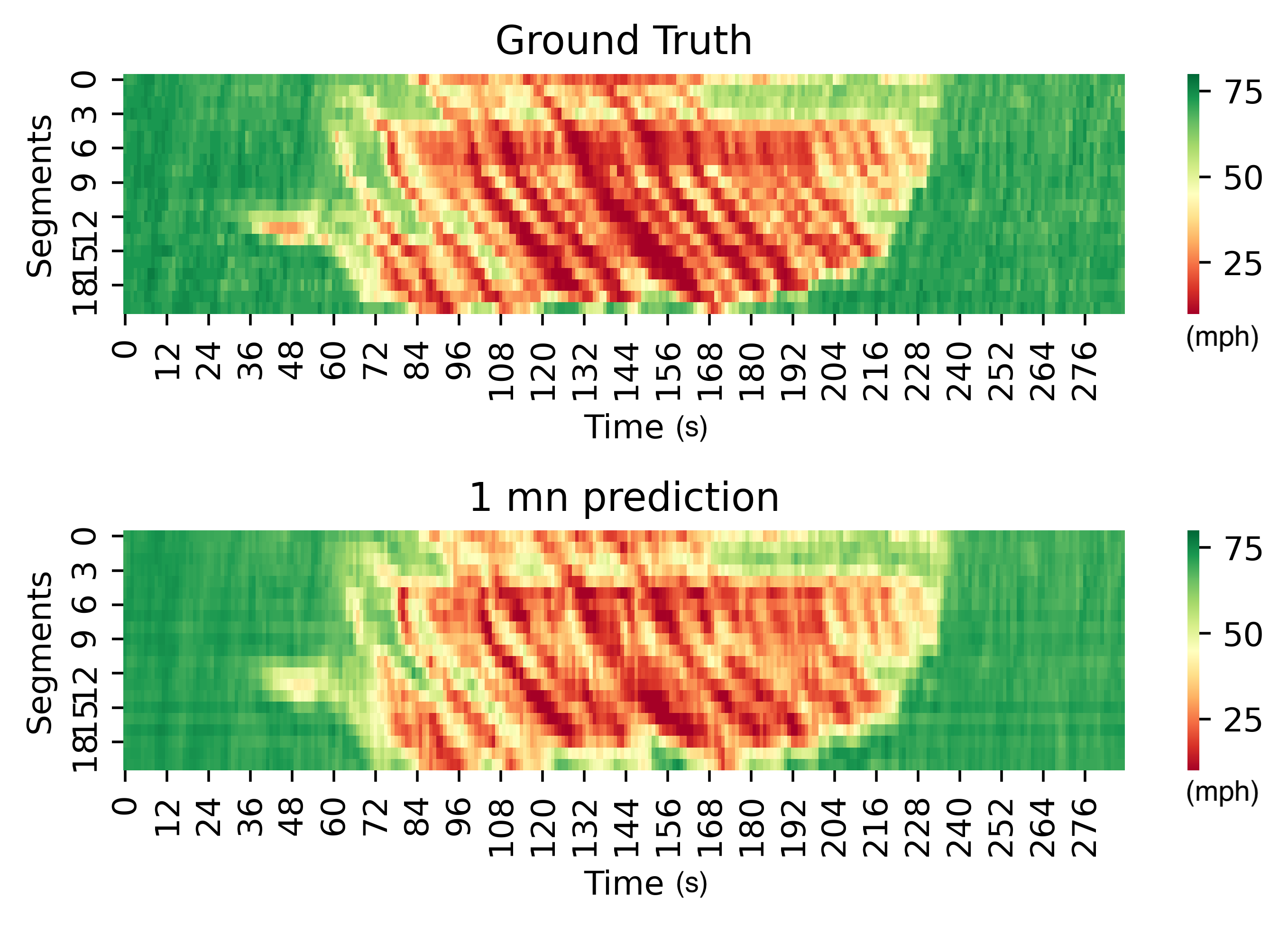}
    \caption{Comparison of heatmap generated from recursive $\mbox{Lap}_3$ SA-LSTM three-minute traffic prediction with a heatmap generated from ground truth data. We observe some blurr in the figure. This is due to the loss of accuracy caused by accumulation error.}
    \label{fig:recursive_heatmap}
\end{figure}

\newpage

\subsection{Velocity curves in diverse stages of traffic}
A more granular and easier to analyse type of representation is the plot of velocity curves in different stages of traffic. Contrarily to heatmaps, a velocity curve focuses on a single timestep and represents the relation between segment and mean velocity of the vehicles in it. This subsection present velocity curves in four representative stages of traffic (free flow of timestep 24, bottleneck of time 48, fully congested on time 180, and dissipation stage of timestep 216) for single step $\mbox{Lap}_3$ SA-LSTM, 3-step $\mbox{Lap}_3$ SA-LSTM, and for the example of a failing case a 3 minute inference via the recursive method with the $\mbox{Lap}_3$ SA-LSTM.

\newpage

\begin{figure*}
\begin{subfigure}[t]{.5\textwidth}
  \centering
    \includegraphics[width=1.0\linewidth]{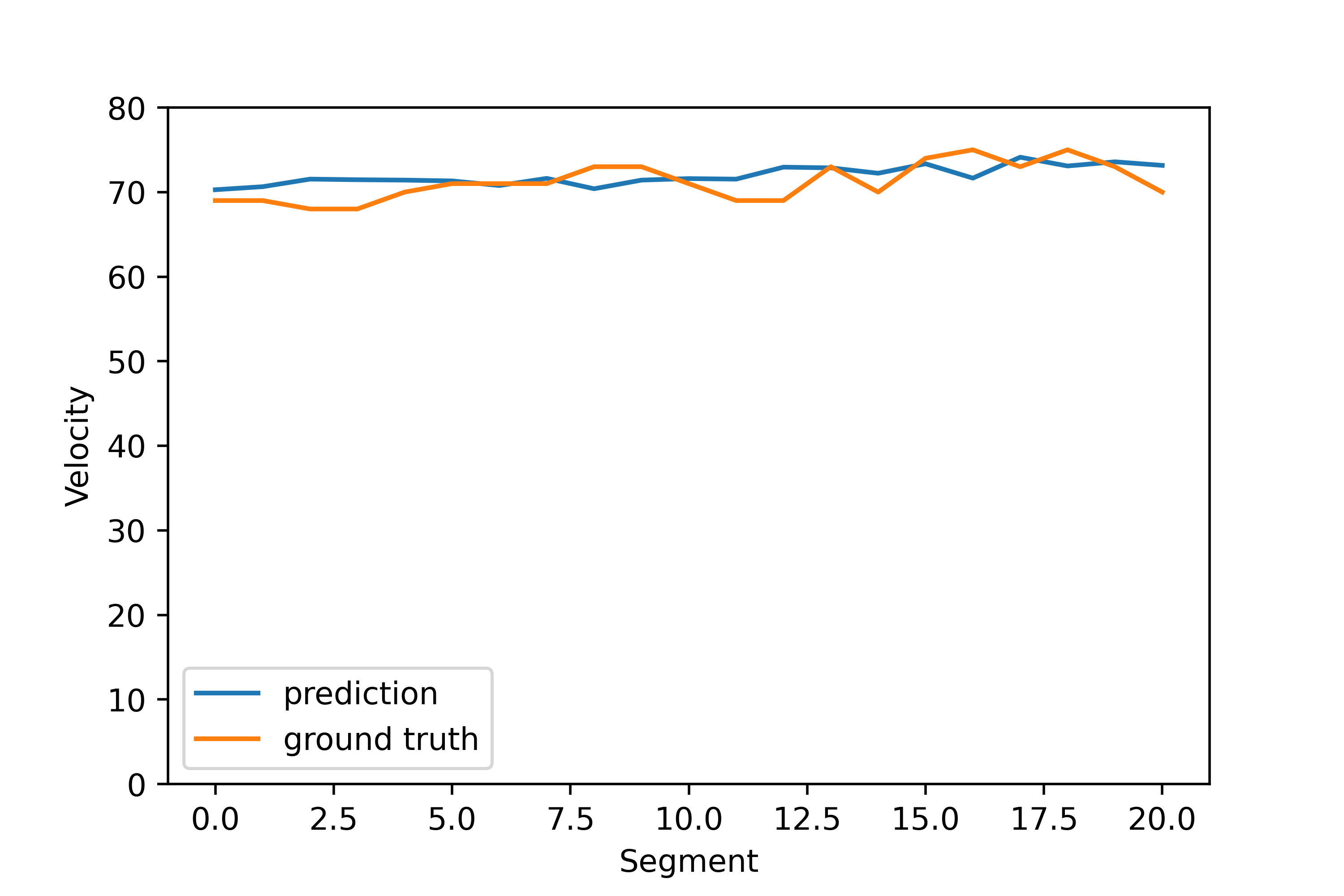}  \caption{Timestep 24: Prediction of free flow stage align with the ground truth around free flow speed.}
\end{subfigure}%
\begin{subfigure}[t]{.5\textwidth}
  \centering
\includegraphics[width=1.0\linewidth]{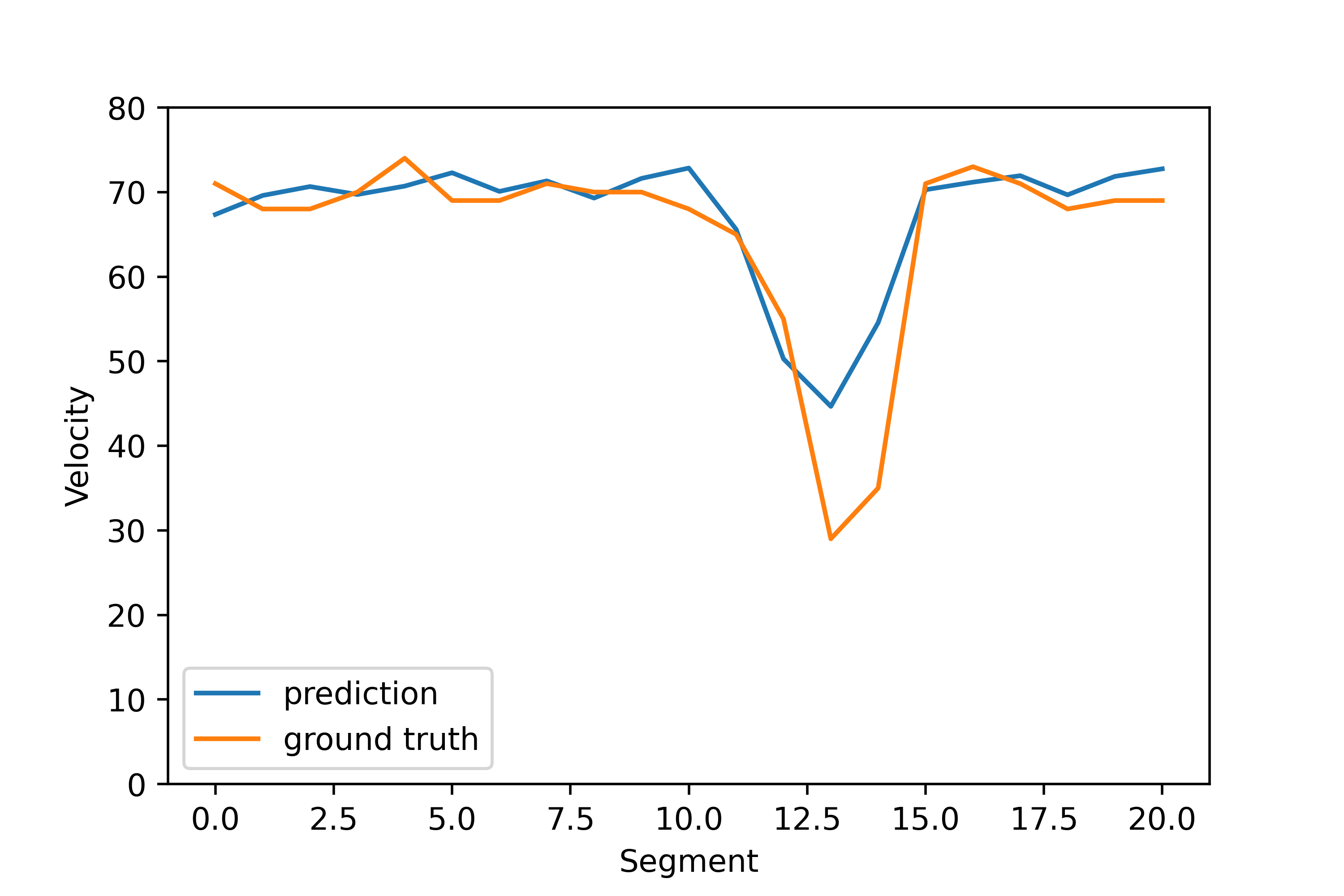}
  \caption{Timestep 48: A bottleneck start to form between segments 11 - 15. The prediction presents the same velocity change pattern with an accurate spatial location of the bottleneck as ground truth. The lowest speed at the bottleneck is underestimated.}
\end{subfigure}
\begin{subfigure}[t]{.5\textwidth}
  \centering
    \includegraphics[width=1.0\linewidth]{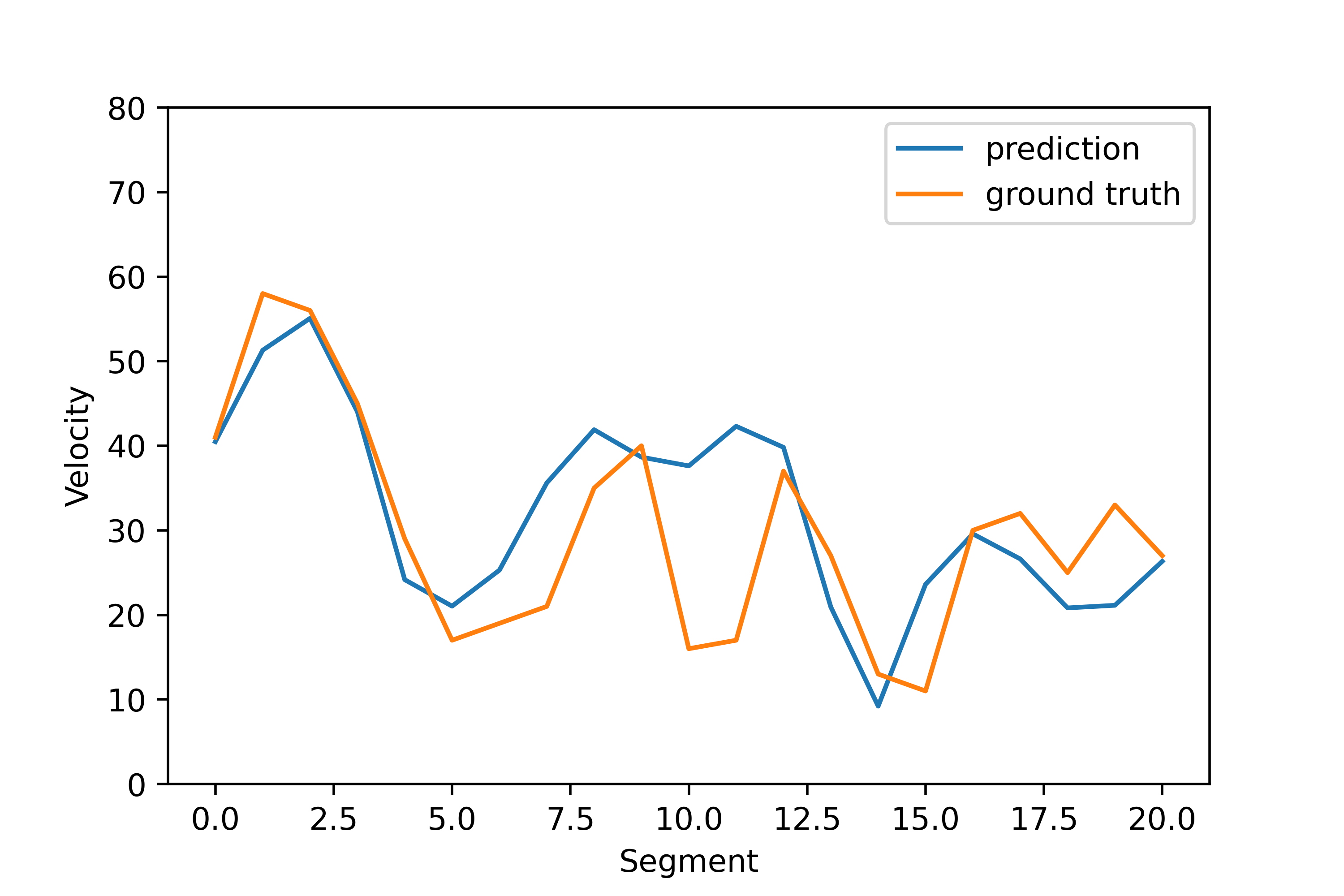}  
    \caption{Timestep 180: During the fully congested stage, the prediction captured the pattern of shockwave, while the prediction of absolute speed value has diversion from the ground truth.}
\end{subfigure}%
\begin{subfigure}[t]{.5\textwidth}
  \centering
    \includegraphics[width=1.0\linewidth]{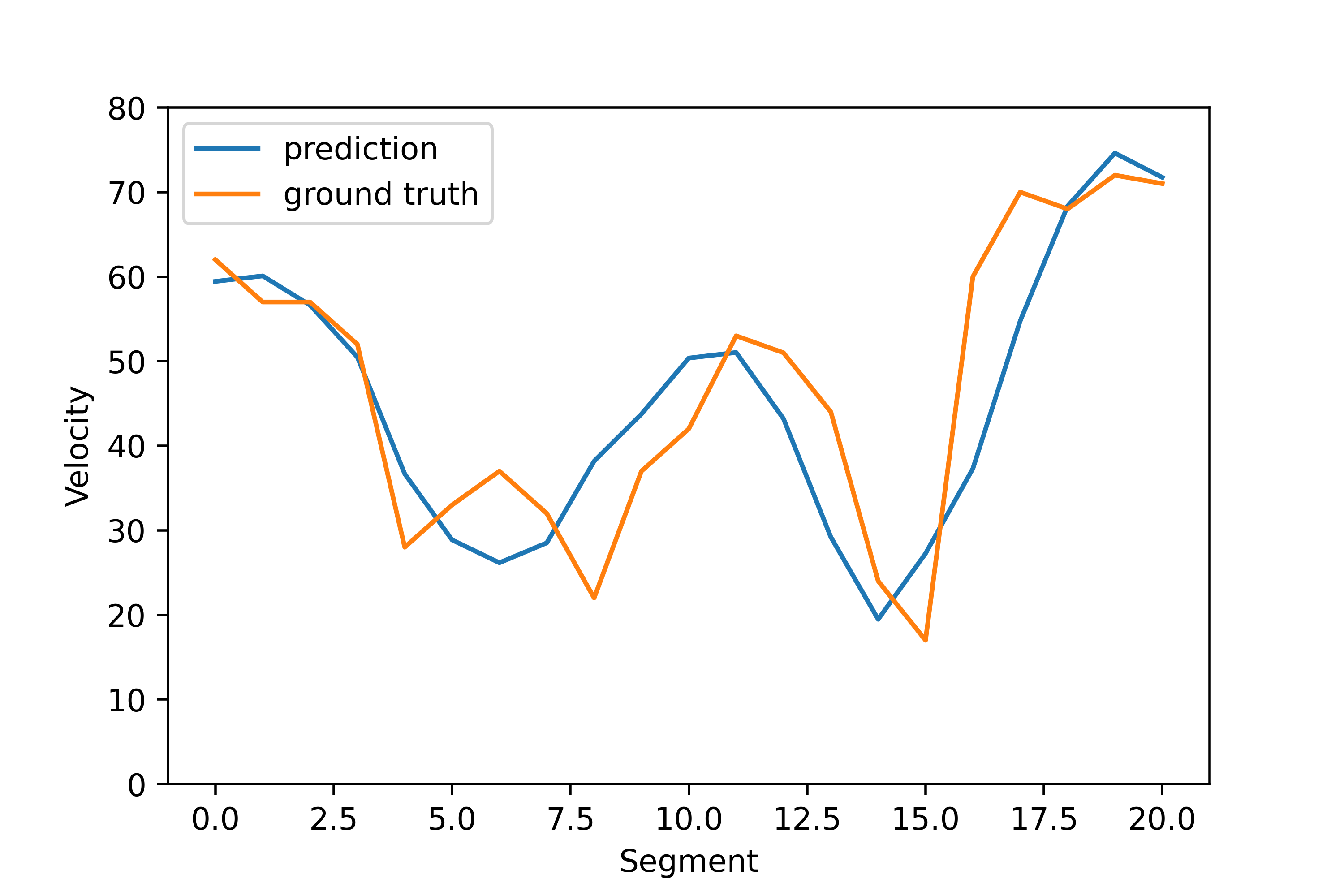}  
    \caption{Timestep 216: During the dissipation stage of the congestion, the prediction is able to capture the speed recovery at the bottleneck and upstream.}
\end{subfigure}%
\caption{Comparison between the ground truth and the three-minute predictions from the recursive $\mbox{Lap}_3$ SA-LSTM during different stages of the congestion lifecycle. Velocities are in mph.}
\label{fig:recursive_plots}
\end{figure*}

\newpage
\clearpage

\section{Conclusion}
This paper tackles the problem of real-time mesoscale traffic forecasting, and presents a fast and accurate method able to extract and analyze both temporal and spatial dependencies in traffic data series. This approach has been analyzed through an extensive ablation study of its components, and compared with state-of-the-art methods for spatio-temporal forecasting to highlight how adapted it is for the studied task. Lastly, we introduced a novel technique for generalization of one-step forecasting method to multi-step forecasting. This method showed to provide the best trade-off inference time on both short-term and long-term forecasting for our considered use case.



\newpage
 \bibliographystyle{elsarticle-num} 
 \bibliography{cas-refs}





\end{document}